\documentclass{article}
\usepackage[margin=1in]{geometry}
\usepackage{graphicx}
\usepackage{multirow}
\usepackage{subcaption}
\usepackage{microtype}
\usepackage{booktabs}
\usepackage{amsmath}
\usepackage{amssymb}
\usepackage{algorithm}
\usepackage{algorithmic}

\usepackage{times}
\usepackage{hyperref}
\usepackage{url}
\usepackage{xcolor}

\title{\textsc{RoboCode}: Learning Reusable Action Primitives for Robot Manipulation}

\author{
Xianyue Peng, Devi Amarsaikhan, Iman Soltani}
\date{}

\begin{document}

\maketitle
\begin{abstract}
Recent robot manipulation policies commonly predict continuous action chunks
directly from observations, without explicitly modeling reusable structure in
the action space. However, manipulation demonstrations often contain recurring
short-horizon motion patterns that are shared across tasks. We introduce
\textsc{RoboCode}, which learns these patterns through a shared soft-gated
primitive action interface. Rather than assigning each action chunk to a single
discrete code, \textsc{RoboCode} represents it as a soft composition of
multiple learned primitives, providing a flexible intermediate representation
of short-horizon actions. The primitive interface is learned from action chunks
pooled across multiple tasks and then frozen while task-specific state encoders
are trained to map current state inputs into the shared latent space.

Experiments on five ManiSkill manipulation tasks address three questions.
First, the learned primitive interface reconstructs action chunks with an
average reconstruction MSE of \(0.0187\). Second, analysis of code activations
reveals structured code reuse across tasks. Third, \textsc{RoboCode} achieves
the highest or tied closed-loop success rate on all evaluated tasks, with an
average success rate of \(75.6\%\), compared with \(68.8\%\) for ACT and
\(41.6\%\) for the larger Diffusion Policy baseline. These results suggest
that soft-gated primitive codebooks provide an effective and interpretable
intermediate action representation for robot manipulation.
\end{abstract}

\section{Introduction}

Robot manipulation is often modeled as a sequence of continuous actions. Across tasks, demonstrations repeatedly contain short motion patterns such as reaching, lifting, placing, and pushing. This suggests that manipulation trajectories may be better understood through reusable primitive structure rather than as independent action chunks.

This raises a question of action representation. Recent token-based robot policies discretize each action dimension into bins~\cite{kim2025openvla}, offering a simple interface but splitting coordinated actions into dimension-wise decisions. Task-level commands and planning operators~\cite{byrnes2025climb} provide semantic structure, but often require a large vocabulary of task-specific actions, limiting generalization across tasks. Primitive-level representations offer a more compositional abstraction, capturing reusable motion units while preserving coordination across action dimensions and time.

Prior work has explored reusable units for robot manipulation~\cite{mao2024dexskills,zhu2025learning,lei2026primaryfine}, including behavior segments, semantic skill spaces, and discrete action modes. However, a single primitive may not be sufficiently expressive for a short action chunk, since the chunk can contain multiple motion factors. Inspired by memory-augmented models~\cite{gong2019memorizing,yang2021memory}, we learn primitive codes as entries in a shared primitive codebook and represent each chunk as a soft composition of multiple entries.

We propose \textsc{RoboCode}, a two-stage framework for learning soft-gated primitive codebooks from robot demonstrations. The first stage jointly learns an action encoder and a shared primitive action
interface from multi-task action chunks. The second stage trains task-specific observation-to-gate policies that use this frozen interface to generate continuous action chunks for execution.

Our aim is to study whether soft-gated primitive codebooks provide an effective intermediate action representation for robot manipulation. The method does not require manually designed primitives, task rewards, or task labels during primitive learning; it discovers reusable action components directly from demonstrations. Our contributions are:

\begin{itemize}
\item We introduce a soft-gated primitive action representation that represents short-horizon manipulation actions as compositions of multiple learned primitives, rather than direct continuous predictions or single-code assignments.

\item The proposed two-stage framework learns a shared primitive action interface from pooled multi-task demonstrations and reuses the frozen interface for task-specific closed-loop control.

\item Experiments on five ManiSkill manipulation tasks demonstrate structured
cross-task primitive reuse. \textsc{RoboCode} achieves the highest or tied
success rate across all evaluated tasks, reaching an average success rate of
\(75.6\%\), compared with \(68.8\%\) for ACT and \(41.6\%\) for DP-Large.
Controlled ablations and transfer experiments further examine the roles of
soft composition and cross-task primitive sharing.

\end{itemize}

\section{Related Work}

\subsection{Imitation Learning for Robot Manipulation}

Imitation learning has become a widely used paradigm for acquiring robot
manipulation behaviors directly from expert demonstrations. Many
behavior-cloning policies formulate control around predicting a current
low-level action from the current observation. Implicit Behavioral Cloning
(IBC)~\cite{florence2022implicit}, for example, models state--action
relationships through an energy-based policy, providing an expressive
alternative to explicit action regression. Perceiver-Actor
(PerAct)~\cite{shridhar2023perceiver} predicts the next discretized 6-DoF
action from language-conditioned 3D observations for multi-task manipulation.
BC-Z~\cite{jang2022bc} extends behavior cloning to large collections of tasks
through language- and video-conditioned policies; its closed-loop policy
executes the current predicted action while also incorporating future
open-loop trajectory prediction during training.

More recent methods increasingly model temporally extended actions directly.
Action Chunking with Transformers (ACT)~\cite{zhao2023learning} predicts
sequences of future actions as chunks, reducing the effective prediction
horizon for fine-grained manipulation. Diffusion Policy
~\cite{chi2025diffusion} models action sequences through conditional diffusion
and supports receding-horizon execution. Temporally extended action prediction
has further scaled to generalist robot policies: Octo~\cite{team2024octo}
learns from diverse robot embodiments and tasks, while RDT-1B
~\cite{liu2025rdt} employs a diffusion Transformer to model complex action
distributions for bimanual manipulation.

Together, these developments expand imitation learning from current-action
prediction toward joint modeling of temporally extended robot behavior.
Beyond predicting such behaviors, a related line of work investigates whether
recurring structure within manipulation trajectories can be captured and
reused through motion primitives and skill abstractions.

\subsection{Motion Primitives and Skill Learning}

Motion primitives and skill abstractions organize robot behavior into reusable
units between task objectives and low-level control. Hierarchical manipulation
approaches decompose multi-phase tasks into reusable skills and learn their
selection and adaptation across different phases
~\cite{kroemer2015towards}. Other methods define primitives around physical
interaction structure. Tactile Dexterity~\cite{hogan2020tactile} combines
manipulation primitives with tactile feedback for contact-rich manipulation,
while HACMan++~\cite{jiang2024hacman++} introduces spatially grounded
parameterized motion primitives that combine discrete primitive selection with
continuous grounding and execution parameters. Motion primitives have also
been integrated with planning and reinforcement learning
~\cite{kicki2024bridging}.

Recent approaches increasingly learn reusable behavioral units directly from
demonstrations or experience. DexSkills~\cite{mao2024dexskills} segments
long-horizon manipulation demonstrations into skills using multimodal sensory
information. XSkill~\cite{xu2023xskill} discovers cross-embodiment skill
prototypes from unlabeled human and robot manipulation demonstrations and
transfers the learned representations to robot control. Mao et al.
~\cite{mao2023learning} learn reusable manipulation mechanisms from
demonstrations and self-play that can be compositionally integrated into
task-and-motion planning. Recent methods further explore semantic and
hierarchical manipulation abstractions
~\cite{zhu2025learning,liu2025himacon} and object-centric interaction
primitives~\cite{pan2025omnimanip}.

These methods differ substantially in what constitutes a reusable behavioral
unit, ranging from parameterized motion primitives to segmented, semantic, and
cross-embodiment skills. They nevertheless share the broader objective of
capturing recurring structure that can be reused across manipulation
behaviors. A further question is how such reusable structure participates
directly in policy learning and action generation.

\subsection{Structured Action Representations and Codebooks}

Recent approaches increasingly incorporate structure into the intermediate
representations through which policies generate robot actions. OpenVLA
~\cite{kim2024openvla} discretizes individual continuous action dimensions
into tokens for autoregressive prediction. FAST~\cite{pertsch2025fast}
instead encodes temporally extended action sequences into compact discrete
tokens through frequency-domain transformation and compression. PF-DAG
~\cite{lei2026primaryfine} separates coarse and fine action generation by
representing action chunks through discrete coarse modes followed by
mode-conditioned continuous action generation.

Learned codebooks provide another mechanism for placing reusable structure
within the action-generation process. STAR~\cite{li2025star} learns reusable
robot skill abstractions through residual vector quantization and models
dependencies among the resulting skill representations for action generation.
In these approaches, structured representations serve not only as descriptions
of robot behavior, but also as intermediate variables through which low-level
actions or action sequences are generated.

\textsc{RoboCode} similarly places reusable structure within the
action-generation interface, but adopts a continuous soft-compositional
representation. Inspired by memory-augmented representation models
~\cite{gong2019memorizing,yang2021memory}, we treat learned codebook entries as
primitive codes and use similarity-based gates to activate multiple entries
simultaneously. Their weighted combination forms a continuous latent
representation that is decoded into an action chunk. As illustrated in
Fig.~\ref{fig:action_representation}, the primitive codebook therefore acts as
a shared intermediate action interface between observation-conditioned policy
learning and continuous action-chunk generation.

\section{Problem Setup}
\label{sec:problem_setup}

We consider multi-task offline imitation learning for robot manipulation.
For each task \(m\), successful demonstrations contain state inputs
\(\mathbf{x}_t^m\) and low-level actions \(\mathbf{a}_t^m\). We construct
short-horizon action chunks
\begin{equation}
    \mathbf{u}_t^m
    =
    \mathbf{a}_{t:t+H-1}^m,
\end{equation}
where \(H\) denotes the action-chunk horizon. The objective is to predict a
future action chunk from the current state input.

Rather than directly regressing \(\mathbf{u}_t^m\) from
\(\mathbf{x}_t^m\), we introduce a shared primitive action interface based on
a learnable primitive codebook
\begin{equation}
    \mathbf{E}
    =
    \{\mathbf{e}_k\}_{k=1}^{K},
    \qquad
    \mathbf{e}_k\in\mathbb{R}^{C},
\end{equation}
where \(K\) is the number of primitive codes and \(C\) is the latent code
dimension. The interface represents an action chunk as a soft composition of
primitive codes through similarity-based gating.

In Stage 1, action chunks pooled across all tasks are used to train an action
encoder \(\operatorname{Enc}\) together with the shared primitive interface. Because Stage 1 pools action chunks across tasks, we omit the task index \(m\)
for notational simplicity:
\begin{equation}
    \mathbf{u}_t
    \xrightarrow{\operatorname{Enc}}
    \mathbf{z}_t
    \xrightarrow{\Gamma(\,\cdot\,;\mathbf{E})}
    \mathbf{g}_t
    \xrightarrow{D(\,\cdot\,;\mathbf{E})}
    \tilde{\mathbf{u}}_t.
\end{equation}
Here, \(\operatorname{Enc}\) maps an action chunk to a latent representation
\(\mathbf{z}_t\), the similarity-based gating operator
\(\Gamma(\,\cdot\,;\mathbf{E})\) produces a soft primitive-gate vector
\(\mathbf{g}_t\), and the action decoder \(D(\,\cdot\,;\mathbf{E})\)
reconstructs the action chunk as \(\tilde{\mathbf{u}}_t\). The primitive
codebook, gating operator, and action decoder are shared across all tasks.

In Stage 2, the learned primitive interface is frozen, and a task-specific
state encoder \(\operatorname{Enc}^m\) is trained for each task:
\begin{equation}
    \mathbf{x}_t^m
    \xrightarrow{\operatorname{Enc}^m}
    \hat{\mathbf{z}}_t^m
    \xrightarrow{\Gamma(\,\cdot\,;\mathbf{E})}
    \hat{\mathbf{g}}_t^m
    \xrightarrow{D(\,\cdot\,;\mathbf{E})}
    \hat{\mathbf{u}}_t^m.
\end{equation}
The task-specific state encoder maps the current state input into the shared
latent space learned in Stage 1. The frozen primitive interface then converts
this latent representation into predicted primitive gates and decodes them
into a future action chunk. This separates learning a shared primitive action
representation from learning task-specific primitive selection policies.
\section{Method}
\label{sec:method}

\begin{figure*}[t]
    \centering
    \includegraphics[
        width=\textwidth
    ]{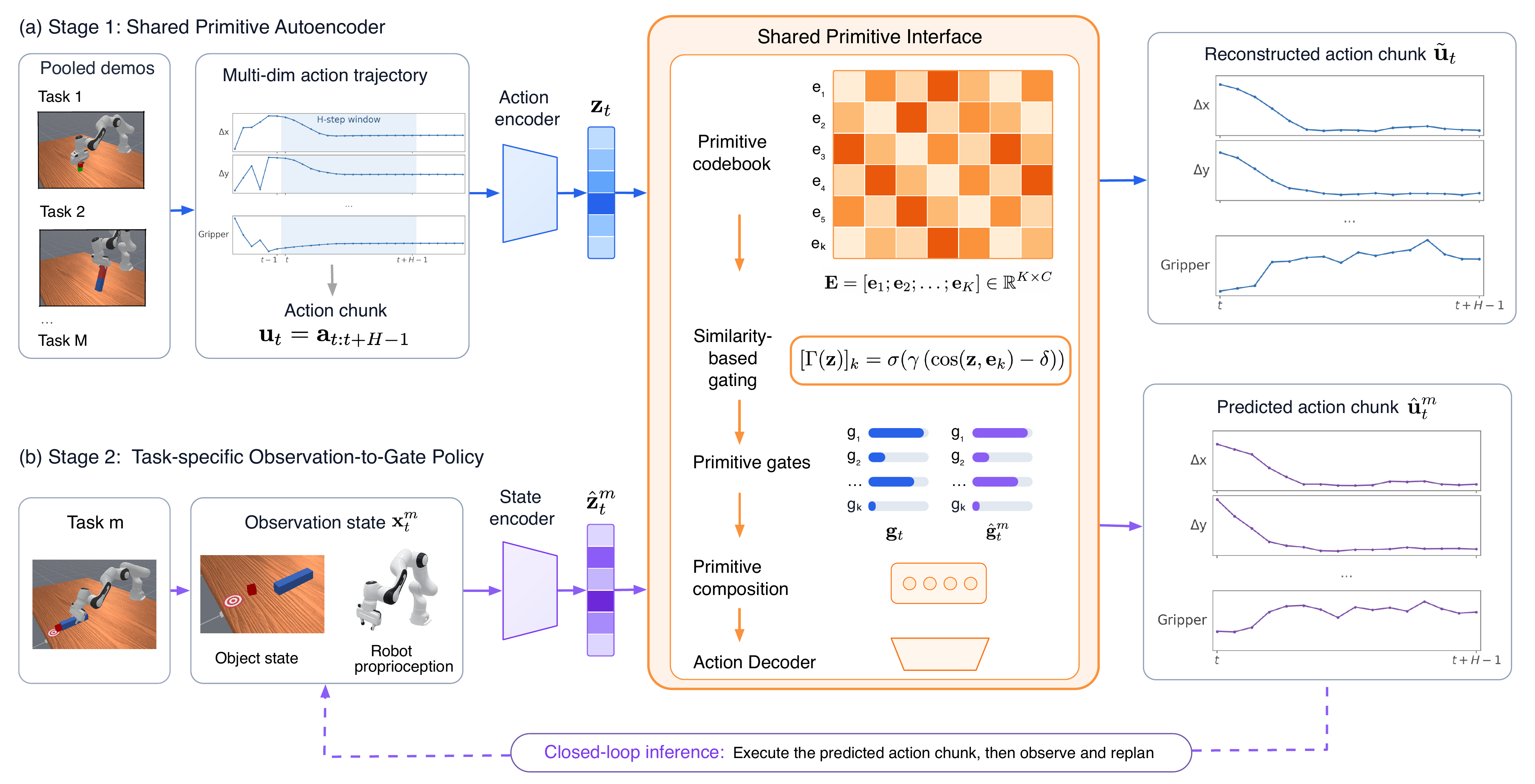}
    \caption{
    Overview of the \textsc{RoboCode} framework.
    Stage 1 learns an action encoder and a shared primitive action interface
    from pooled multi-task action chunks.
    Stage 2 trains task-specific state encoders to map observations into the
    shared latent space while keeping the primitive interface frozen.
    At inference time, predicted action chunks are executed in a
    receding-horizon manner.
    }
    \label{fig:framework}
\end{figure*}

\subsection{Stage 1: Shared Primitive Autoencoder}
\label{subsec:stage1}

Stage 1 trains an action encoder together with the shared primitive action
interface introduced in Sec.~\ref{sec:problem_setup}. The action encoder maps
each action chunk into the latent space of the primitive codebook, while the
shared interface converts the resulting latent representation into soft
primitive gates and reconstructs the original action chunk.

\paragraph{Action encoding.}
Given a normalized action chunk \(\mathbf{u}_t\), the action encoder
\(\operatorname{Enc}\) produces a chunk-level latent representation:
\begin{equation}
    \mathbf{z}_t
    =
    \operatorname{Enc}(\mathbf{u}_t).
\end{equation}

\paragraph{Similarity-based primitive gating.}
The latent representation is compared with each primitive code using cosine
similarity, followed by a shifted sigmoid:
\begin{equation}
    \mathbf{g}_t
    =
    \Gamma(\mathbf{z}_t;\mathbf{E}),
    \qquad
    [\Gamma(\mathbf{z};\mathbf{E})]_k
    =
    \sigma\!\left(
        \gamma\left(
            \cos(\mathbf{z},\mathbf{e}_k)-\delta
        \right)
    \right).
\end{equation}
Here, \(\Gamma\) denotes the similarity-based gating operator,
\(\sigma(\cdot)\) is the sigmoid function, \(\gamma\) controls the gate
sharpness, and \(\delta\) is the similarity threshold. Unlike hard vector
quantization, the soft gates allow one action chunk to activate multiple
primitive codes and be represented as their composition.

\paragraph{Primitive composition and action decoding.}
Given the soft gate vector \(\mathbf{g}_t\), the primitive codes are combined
into a gated latent representation:
\begin{equation}
    \mathbf{c}_t
    =
    \sum_{k=1}^{K}
    g_{t,k}\mathbf{e}_k.
\end{equation}
The shared action decoder then reconstructs the action chunk:
\begin{equation}
    \tilde{\mathbf{u}}_t
    =
    D(\mathbf{g}_t;\mathbf{E})
    =
    d_\phi(\mathbf{c}_t).
\end{equation}
The reconstructed chunk \(\tilde{\mathbf{u}}_t\) approximates the input action
chunk through the shared primitive action interface.

\paragraph{Autoencoding objective.}
We train the primitive autoencoder with
\begin{equation}
    \mathcal{L}_{AE}
    =
    \lambda_{\mathrm{rec}}\mathcal{L}_{\mathrm{rec}}
    +
    \lambda_{\mathrm{proj}}\mathcal{L}_{\mathrm{proj}}
    +
    \lambda_{\mathrm{sparse}}\mathcal{L}_{\mathrm{sparse}}
    +
    \lambda_{\mathrm{div}}\mathcal{L}_{\mathrm{div}},
\end{equation}
where the loss terms are
\begin{equation}
\begin{aligned}
    \mathcal{L}_{\mathrm{rec}}
    &=
    \mathrm{MSE}
    \left(
        \tilde{\mathbf{u}}_t,
        \mathbf{u}_t
    \right), \\
    \mathcal{L}_{\mathrm{proj}}
    &=
    \mathrm{MSE}
    \left(
        \mathbf{c}_t,
        \mathbf{z}_t
    \right), \\
    \mathcal{L}_{\mathrm{sparse}}
    &=
    \sum_{k=1}^{K}
    g_{t,k}, \\
    \mathcal{L}_{\mathrm{div}}
    &=
    \frac{1}{K(K-1)}
    \sum_{i\neq j}
    \cos(\mathbf{e}_i,\mathbf{e}_j)^2.
\end{aligned}
\end{equation}
The reconstruction and projection terms preserve chunk-level reconstruction
and latent consistency, while the sparsity term regularizes gate usage and
the diversity term encourages codebook diversity.

\subsection{Stage 2: Observation-to-Gate Policy}
\label{subsec:stage2}

Stage 2 learns task-specific primitive selection from the current state input.
After Stage 1, the primitive codebook, similarity-based gating operator, and
action decoder are frozen. For each task \(m\), we train a task-specific state
encoder \(\operatorname{Enc}^m\):
\begin{equation}
    \hat{\mathbf{z}}_t^m
    =
    \operatorname{Enc}^m(\mathbf{x}_t^m).
\end{equation}
The resulting latent representation is passed through the frozen primitive
interface:
\begin{equation}
    \hat{\mathbf{g}}_t^m
    =
    \Gamma(\hat{\mathbf{z}}_t^m;\mathbf{E}),
    \qquad
    \hat{\mathbf{u}}_t^m
    =
    D(\hat{\mathbf{g}}_t^m;\mathbf{E}).
\end{equation}
In our implementation, \(\mathbf{x}_t^m\) is constructed from task-specific
object and goal states together with robot proprioceptive states.

We train \(\operatorname{Enc}^m\) by matching the predicted gates to the target
gates inferred from expert future action chunks, with an additional action
consistency term:
\begin{equation}
    \mathcal{L}_{\mathrm{policy}}
    =
    \mathrm{MSE}\!\left(
        \hat{\mathbf{g}}_t^m,
        \Gamma\!\left(
            \operatorname{Enc}(\mathbf{u}_t^m);
            \mathbf{E}
        \right)
    \right)
    +
    \alpha\,
    \mathrm{MSE}\!\left(
        \hat{\mathbf{u}}_t^m,
        \mathbf{u}_t^m
    \right).
\end{equation}

\subsection{Closed-Loop Execution}
\label{subsec:closed_loop}

During inference, \textsc{RoboCode} operates in a receding-horizon manner. Given the current state input \(\mathbf{x}_t^m\), the task-specific
state encoder produces \(\hat{\mathbf{z}}_t^m\), and the frozen primitive
interface converts it into a predicted action chunk
\(\hat{\mathbf{u}}_t^m\). The controller executes only the first part of the
length-\(H\) chunk before observing the environment again and replanning. This
allows temporally extended action chunks to be executed with closed-loop
feedback.

\section{Experiments}
\label{sec:experiments}
We evaluate \textsc{RoboCode} on multi-task robot manipulation. Specifically, we ask three questions: (i) whether action chunks can be reconstructed through a shared primitive codebook, (ii) whether the learned codes capture reusable motion patterns across tasks, and (iii) whether the resulting policies improve closed-loop control.

\subsection{Experimental Setup}
\label{subsec:exp_setup}

\paragraph{Dataset.}
We conduct experiments on a multi-task demonstration dataset collected from
five ManiSkill manipulation tasks: LiftPegUpright, PickCube, PokeCube,
PushCube, and StackCube. For each task, we use successful demonstrations and
construct training samples by applying a sliding window over each trajectory.
At time \(t\), each sample consists of a short-horizon action chunk
\(\mathbf{u}_t^m=\mathbf{a}_{t:t+H-1}^m\) and a state input
\(\mathbf{x}_t^m=[\mathbf{o}_t^m;\mathbf{r}_t^m]\).

Each low-level action follows ManiSkill's seven-dimensional
end-effector delta-pose control mode
(\texttt{pd\_ee\_delta\_pose}), consisting of incremental
end-effector translations
\((\Delta x,\Delta y,\Delta z)\), incremental rotational components
\((\Delta r_x,\Delta r_y,\Delta r_z)\), and a gripper command.

\begin{figure*}[t]
    \centering

    \begin{subfigure}[t]{0.19\textwidth}
        \centering
        \includegraphics[width=\linewidth]
        {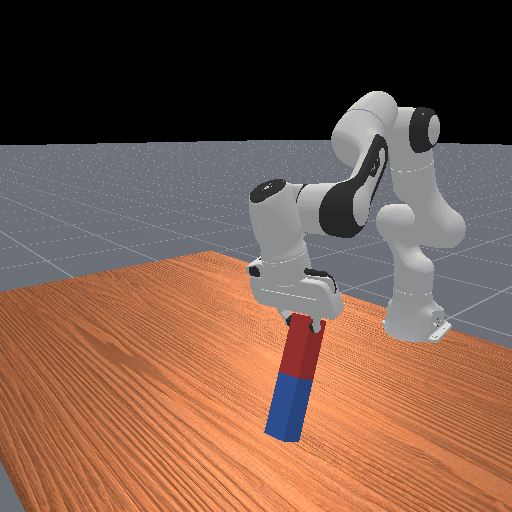}
        \caption{LiftPegUpright}
    \end{subfigure}
    \hfill
    \begin{subfigure}[t]{0.19\textwidth}
        \centering
        \includegraphics[width=\linewidth]
        {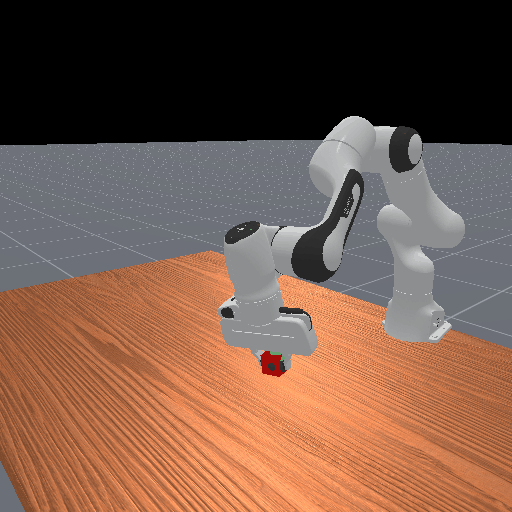}
        \caption{PickCube}
    \end{subfigure}
    \hfill
    \begin{subfigure}[t]{0.19\textwidth}
        \centering
        \includegraphics[width=\linewidth]
        {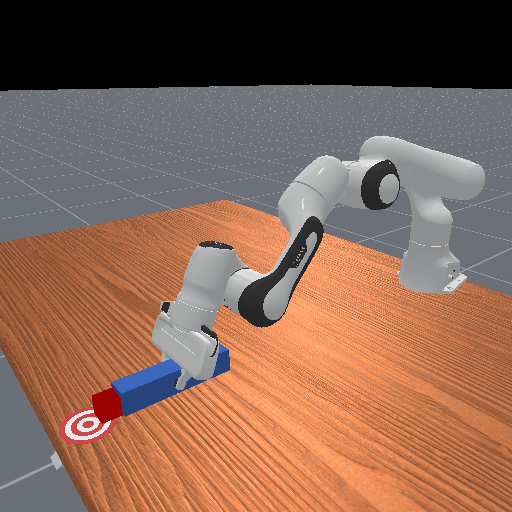}
        \caption{PokeCube}
    \end{subfigure}
    \hfill
    \begin{subfigure}[t]{0.19\textwidth}
        \centering
        \includegraphics[width=\linewidth]
        {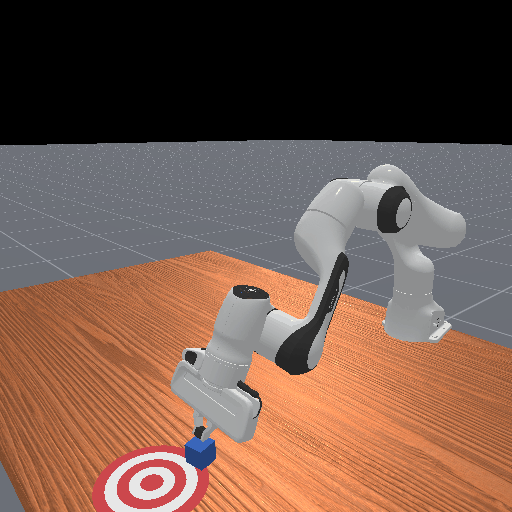}
        \caption{PushCube}
    \end{subfigure}
    \hfill
    \begin{subfigure}[t]{0.19\textwidth}
        \centering
        \includegraphics[width=\linewidth]
        {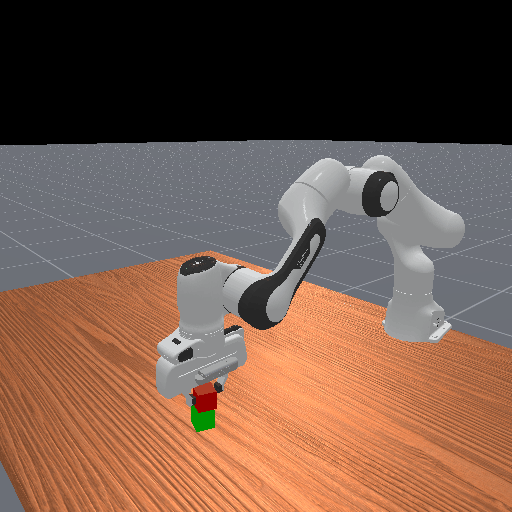}
        \caption{StackCube}
    \end{subfigure}

    \caption{
    Representative environments of the five ManiSkill manipulation tasks
    used in our experiments.
    }
    \label{fig:tasks}
\end{figure*}

The state component \(\mathbf{o}_t^m\) contains the task-relevant object
poses and goal positions. For PickCube, PokeCube, PushCube, and StackCube,
it additionally includes the gripper width and a three-dimensional relative
position. The relative position is computed as the goal position minus the
cube position for PickCube, PokeCube, and PushCube, and as the position of
cube B minus that of cube A for StackCube. The robot state
\(\mathbf{r}_t^m\) is an 18-dimensional proprioceptive vector formed by
concatenating nine joint positions and nine joint velocities.
Fig.~\ref{fig:tasks} shows representative snapshots of the five task
environments, and Table~\ref{tab:dataset} summarizes the dataset statistics.

\begin{table}[t]
\centering
\caption{Dataset statistics for the five ManiSkill tasks.}
\label{tab:dataset}
\begin{tabular}{llrrr}
\toprule
Task & \(\mathbf{o}_t^m\) contents & Demos & Chunks & Input dim \\
\midrule
LiftPegUpright
& peg pose
& 1015 & 7142 & 25 \\

PickCube
& cube pose, goal, width, relative position
& 1013 & 9046 & 32 \\

PokeCube
& cube pose, goal, width, relative position
& 579 & 3194 & 32 \\

PushCube
& cube pose, goal, width, relative position
& 1018 & 4766 & 32 \\

StackCube
& two cube poses, width, relative position
& 995 & 6462 & 36 \\
\midrule
Total
& --
& 4620 & 30610 & -- \\
\bottomrule
\end{tabular}
\end{table}

\paragraph{Model Configuration.}
We use chunk length \(H=16\), window stride \(s=4\), codebook size \(K=32\),
code dimension \(C=64\), and batch size 256. Stage 1 trains the shared
primitive autoencoder on action chunks pooled across all tasks with action
normalization for 400 epochs. Stage 2 freezes the learned primitive interface
and trains a task-specific state encoder \(\operatorname{Enc}^m\) for each task
for 200 epochs. During closed-loop evaluation, the model replans every 8 steps.

\paragraph{Baselines.}
For closed-loop evaluation, we compare against ACT and two Diffusion Policy
variants, denoted Diffusion Policy and DP-Large. To ensure a controlled
comparison, all methods are trained on the same processed demonstrations and
use the same state input
\(\mathbf{x}_t^m=[\mathbf{o}_t^m;\mathbf{r}_t^m]\).
ACT, Diffusion Policy, and \textsc{RoboCode} have comparable parameter counts,
whereas DP-Large is substantially larger. Table~\ref{tab:model_size} reports
the parameter counts of all methods and the two stages of \textsc{RoboCode}.

\begin{table*}[t]
\centering
\caption{
Model sizes in millions of parameters. \textsc{RoboCode} Stage~1 is shared
across tasks and is therefore counted once in the total, while Stage~2 is
task specific.
}
\label{tab:model_size}
\resizebox{\textwidth}{!}{
\begin{tabular}{lcccccc}
\toprule
Method or Component
& LiftPegUpright
& PickCube
& PokeCube
& PushCube
& StackCube
& Total \\
\midrule
ACT
& 0.270 & 0.271 & 0.271 & 0.271 & 0.271 & 1.353 \\
Diffusion Policy
& 0.313 & 0.323 & 0.323 & 0.323 & 0.328 & 1.609 \\
DP-Large
& 4.270 & 4.320 & 4.320 & 4.320 & 4.348 & 21.578 \\
\midrule
\textsc{RoboCode} Stage~1
& 0.148 & 0.148 & 0.148 & 0.148 & 0.148 & 0.148 \\
\textsc{RoboCode} Stage~2
& 0.089 & 0.091 & 0.091 & 0.091 & 0.092 & 0.453 \\
\textsc{RoboCode} Total
& \textbf{0.237} & \textbf{0.239} & \textbf{0.239}
& \textbf{0.239} & \textbf{0.240} & \textbf{0.601} \\
\bottomrule
\end{tabular}
}
\end{table*}

\subsection{Training Diagnostics and Primitive Analysis}
\label{subsec:training_primitive_analysis}

\begin{figure*}[t]
    \centering

    \begin{subfigure}[t]{0.48\textwidth}
        \centering
        \includegraphics[width=\linewidth]{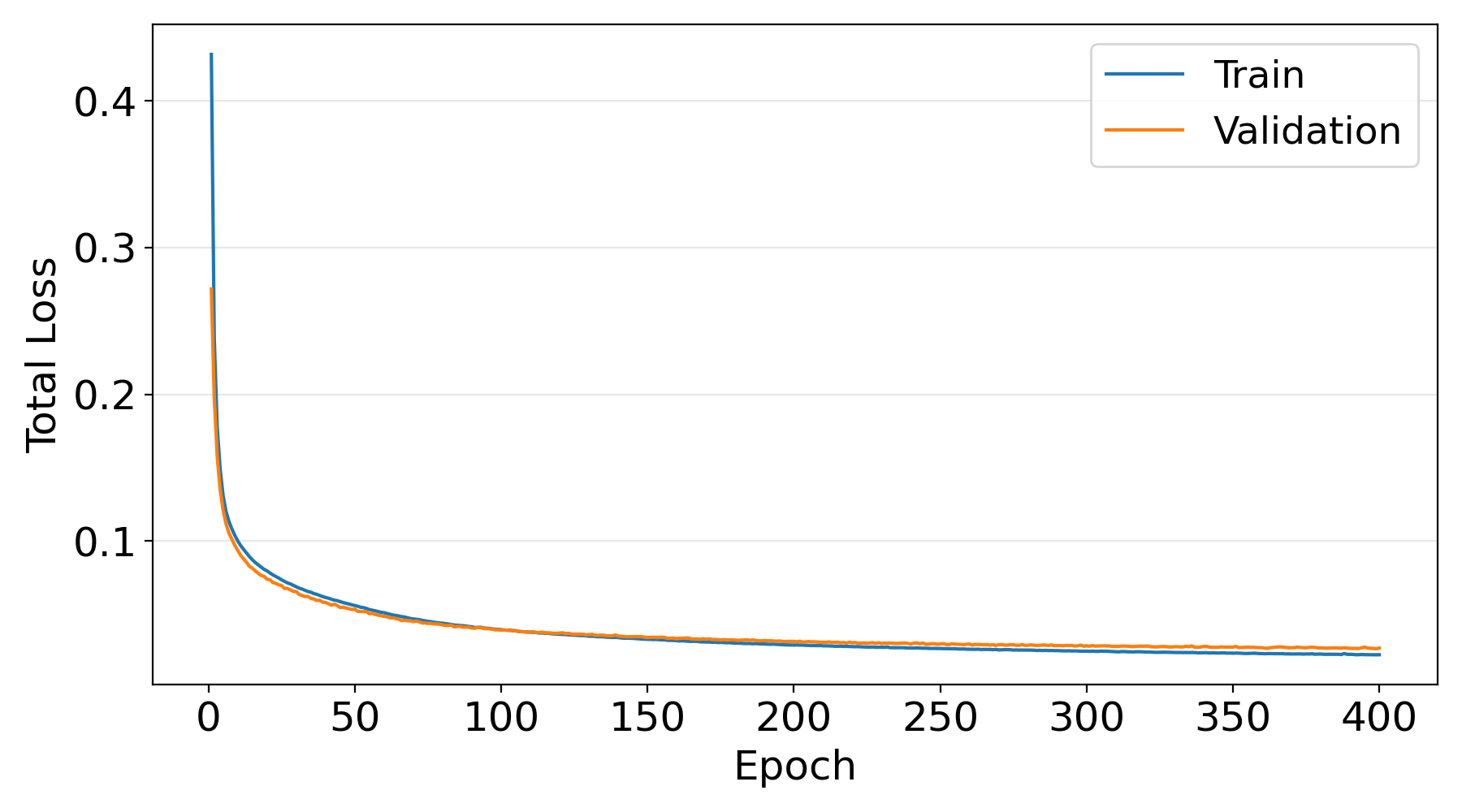}
        \caption{Stage 1 autoencoder training}
        \label{fig:autoencoder_training}
    \end{subfigure}
    \hfill
    \begin{subfigure}[t]{0.48\textwidth}
        \centering
        \includegraphics[width=\linewidth]{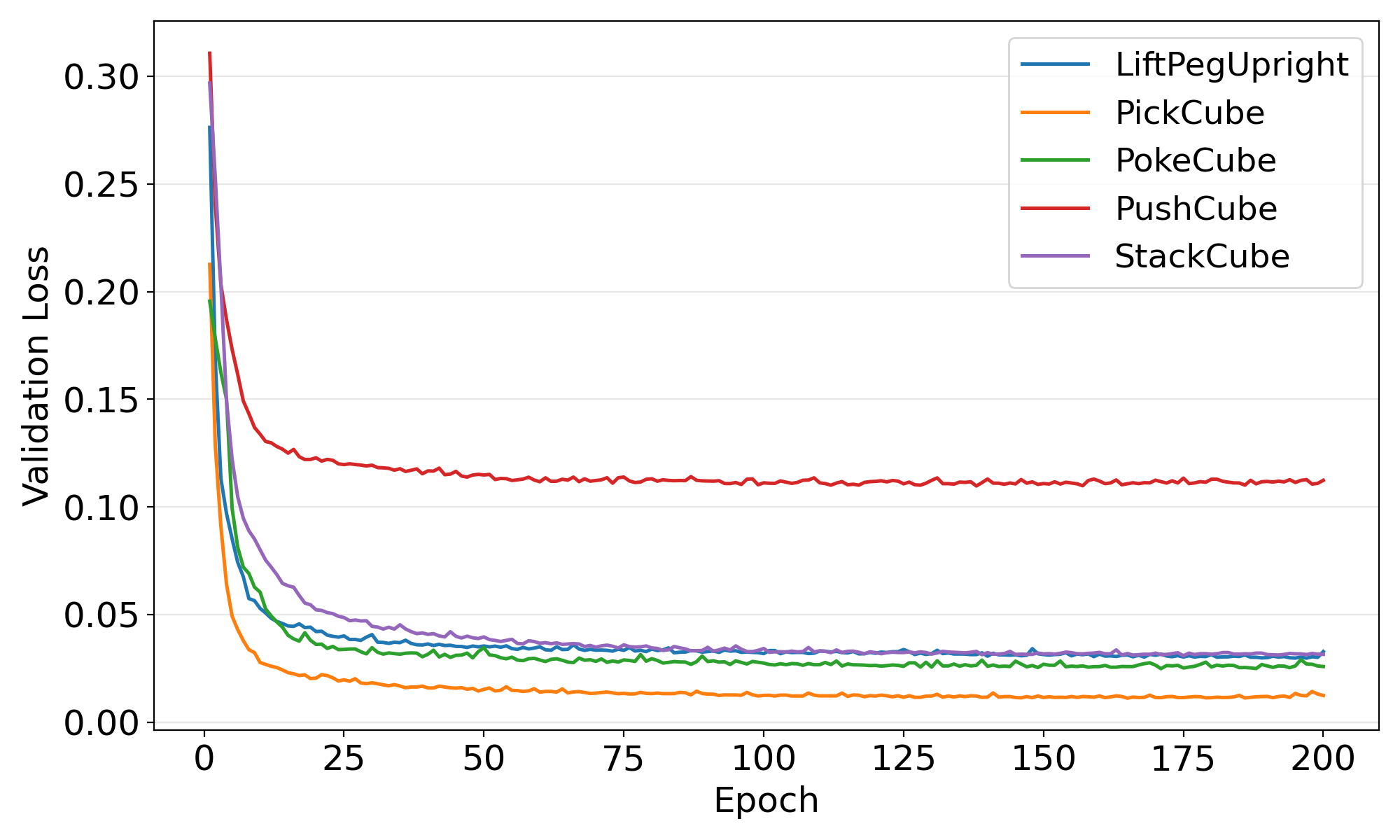}
        \caption{Stage 2 policy training}
        \label{fig:policy_training}
    \end{subfigure}

    \caption{
    Training diagnostics for the two-stage framework.
    (a) Training and validation losses of the shared primitive autoencoder.
    (b) Validation losses of the task-specific observation-to-gate policies.
    }
    \label{fig:training_curves}
\end{figure*}

\paragraph{Convergence.}
Fig.~\ref{fig:training_curves} presents the optimization behavior of the two
training stages. In Stage 1, both the training and validation losses decrease
rapidly during the initial epochs and then improve gradually throughout
training. The two curves remain closely aligned, with no pronounced increase
in validation loss, indicating no substantial overfitting. In Stage 2, the
validation losses decrease sharply during the first 10--20 epochs and
subsequently converge to stable, task-dependent levels across all five tasks.

\paragraph{Offline Accuracy.}

\begin{table}[t]
\centering
\caption{Offline errors of the shared primitive interface and task-specific policies. Lower is better.}
\label{tab:offline_metrics}
\begin{tabular}{lccc}
\toprule
& \multicolumn{1}{c}{Stage 1}
& \multicolumn{2}{c}{Stage 2} \\
\cmidrule(lr){2-2}
\cmidrule(lr){3-4}
Task
& Reconstruction MSE
& Gate MSE
& Action MSE \\
\midrule
LiftPegUpright & 0.0203 & \(1.35\times10^{-4}\) & 0.0653 \\
PickCube       & 0.0082 & \(4.16\times10^{-5}\) & 0.0247 \\
PokeCube       & 0.0148 & \(6.18\times10^{-5}\) & 0.0516 \\
PushCube       & 0.0378 & \(5.65\times10^{-4}\) & 0.2236 \\
StackCube      & 0.0210 & \(1.19\times10^{-4}\) & 0.0628 \\
\midrule
Average        & 0.0187 & \(1.56\times10^{-4}\) & 0.0733 \\
\bottomrule
\end{tabular}
\end{table}

Table~\ref{tab:offline_metrics} reports three offline metrics. Reconstruction
MSE, \(\mathrm{MSE}(\tilde{\mathbf{u}}_t,\mathbf{u}_t)\), evaluates the
Stage 1 action-chunk reconstruction, while gate MSE,
\(\mathrm{MSE}(\hat{\mathbf{g}}_t^m,
\Gamma(\operatorname{Enc}(\mathbf{u}_t^m);\mathbf{E}))\), and action MSE,
\(\mathrm{MSE}(\hat{\mathbf{u}}_t^m,\mathbf{u}_t^m)\), evaluate the Stage 2
gate and decoded-action predictions, respectively.

The average reconstruction MSE is \(0.0187\), indicating that the learned
primitive interface retains the information needed to reconstruct the
demonstrated action chunks. Gate prediction errors remain small across all
tasks, with an average MSE of \(1.56\times10^{-4}\), while the average action
MSE is \(0.0733\).

PushCube exhibits the largest offline errors, with a reconstruction MSE of
\(0.0378\) and an action MSE of \(0.2236\). As shown later in the closed-loop
evaluation, the relatively high offline action MSE on PushCube does not indicate
poor task performance. The task admits a broad set of successful action
trajectories, so a predicted chunk may differ from the demonstrated expert
chunk while still moving the cube toward the target region.

\paragraph{Primitive Semantics.}
\begin{figure*}[t]
    \centering
    \includegraphics[width=\textwidth]
    {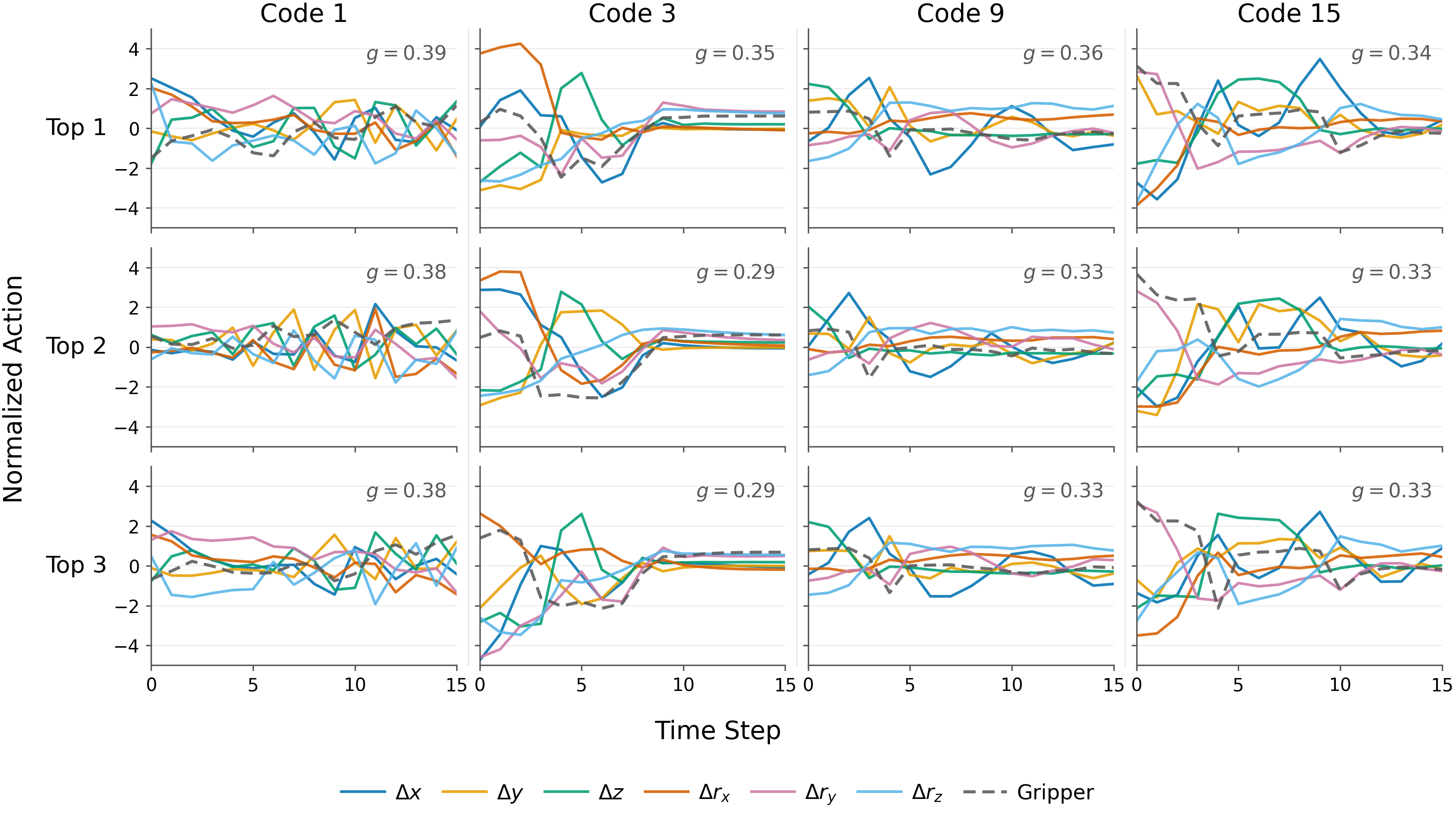}
    \caption{
    Top-activated validation action chunks for four representative primitive
    codes learned in Stage 1. Each column corresponds to one code, and each row
    shows one of its three highest-activation chunks ranked by \(g_{t,k}\).
    The curves represent task-wise standardized seven-dimensional
    end-effector actions over the 16-step action horizon.
    }
    \label{fig:top_chunks}
\end{figure*}

Fig.~\ref{fig:top_chunks} examines the temporal action patterns associated with
selected primitive codes. For each code \(k\), validation chunks are ranked by
the Stage 1 activation \(g_{t,k}\), and the three highest-ranked
examples are shown. Code~3 exhibits a clear two-phase structure, with
substantial multi-channel variation during the first half of the chunk,
followed by nearly constant commands after approximately Steps~8--9.
Code~9 shows a consistent early transition across several channels, with
substantially lower variation during the second half. For Code~15, the three
top-activated chunks exhibit similar recurring multi-channel changes at
comparable positions within the action horizon, with several channels
continuing to vary during the second half. Code~1 is characterized by
sustained, relatively high-frequency fluctuations across multiple action
channels, particularly during the second half of the chunk. Overall, these
within-code regularities indicate that the learned codebook captures distinct
temporal coordination patterns across multiple action dimensions rather than
independently discretizing individual action channels.

\paragraph{Code Usage.}
\begin{figure*}[t]
    \centering
    \includegraphics[
        width=\textwidth
    ]{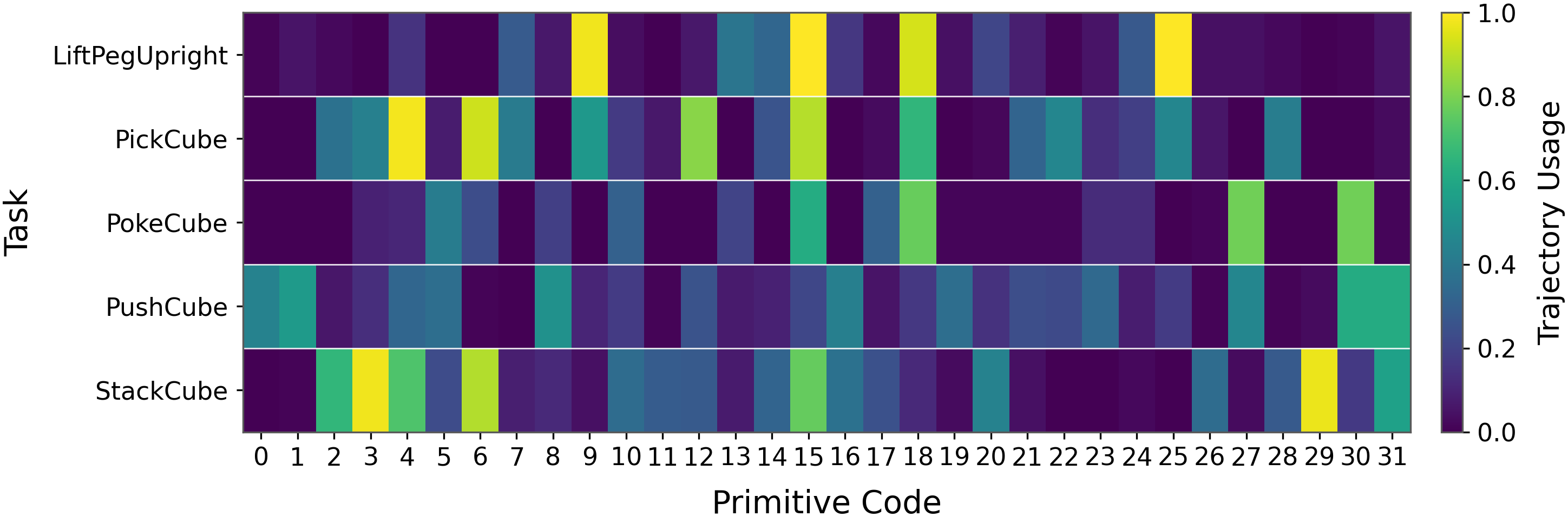}
    \caption{
    Trajectory-level primitive-code usage across the five ManiSkill tasks.
    Each cell reports the fraction of validation trajectories in which the
    corresponding code is activated by at least one action chunk, using a
    gate threshold of \(0.1\).
    }
    \label{fig:task_code_heatmap}
\end{figure*}

Fig.~\ref{fig:task_code_heatmap} reveals both cross-task sharing and
task-dependent specialization in the learned codebook. Because each
manipulation trajectory is divided into multiple short-horizon action chunks,
a single trajectory can invoke several codes at different stages of execution.
Code~15, for example, is frequently used across multiple tasks, suggesting a
broadly reusable motion component. PickCube and StackCube also exhibit similar
high usage of Codes~2, 3, 4, and~6. This overlap may reflect motion components
shared by the two tasks, both of which require grasping and lifting
similarly sized cubes. In contrast, some codes show substantially more
task-specific usage; Code~29, for instance, is activated primarily in
StackCube trajectories. Overall, the codebook exhibits compositional reuse:
tasks draw on overlapping subsets of primitives while retaining specialized
codes for task-dependent behaviors.

\subsection{Closed-Loop Evaluation}
\label{subsec:closed_loop_eval}

We evaluate \textsc{RoboCode}, ACT, Diffusion Policy, and DP-Large in
closed-loop ManiSkill environments. \textsc{RoboCode} follows a
receding-horizon execution scheme. At each decision point, the task-specific
state encoder maps the current state input into the shared latent space, and
the frozen primitive interface converts the resulting representation into a
length-\(H\) action chunk. The controller executes the first eight actions
before observing the environment again and replanning. All methods are
evaluated under the same episode initialization protocol, task-success
criteria, and maximum episode horizon of 50 environment steps.

We report three complementary metrics in
Table~\ref{tab:closed_loop_metrics}. Success rate measures the reliability of
task completion and serves as the primary metric. Average return is computed
over all evaluation episodes and provides an auxiliary indication of partial
task progress, including unsuccessful trials. Because accumulated return
depends on the task reward, episode duration, and termination behavior, it
should not be interpreted as a direct measure of execution efficiency. Steps
to success are therefore reported separately and averaged over successful
episodes only.

\begin{table*}[t]
\centering
\caption{
Closed-loop control performance across the five ManiSkill tasks.
Success rate and average return are computed over all evaluation episodes,
while steps to success are averaged over successful episodes only.
DP-Large increases the U-Net channel dimensions of Diffusion Policy from
\([24,64]\) to \([64,128,256]\).
A dash indicates that no successful episode was observed.
}
\label{tab:closed_loop_metrics}
\resizebox{\textwidth}{!}{
\begin{tabular}{llccccc}
\toprule
Metric & Method
& LiftPegUpright
& PickCube
& PokeCube
& PushCube
& StackCube \\
\midrule

\multirow{4}{*}{Success Rate (\%) $\uparrow$}
& ACT
& 56 & 86 & 50 & \textbf{98} & 54 \\
& Diffusion Policy
& 14 & 22 & 24 & 34 & 0 \\
& DP-Large
& 20 & 56 & 36 & 68 & 28 \\
& \textsc{RoboCode}
& \textbf{80} & \textbf{92} & \textbf{52}
& \textbf{98} & \textbf{56} \\

\midrule

\multirow{4}{*}{Average Return $\uparrow$}
& ACT
& \textbf{15.14} & 6.51 & 6.28 & 1.97 & 7.77 \\
& Diffusion Policy
& 10.93 & 3.02 & 4.91 & \textbf{2.61} & 3.26 \\
& DP-Large
& 10.88 & 5.93 & 4.69 & 2.60 & 5.57 \\
& \textsc{RoboCode}
& 14.47 & \textbf{7.70} & \textbf{7.33}
& 1.98 & \textbf{8.37} \\

\midrule

\multirow{4}{*}{Steps to Success $\downarrow$}
& ACT
& 16.43 & 14.91 & 49.00 & 6.29 & 13.44 \\
& Diffusion Policy
& \textbf{16.29} & 15.00 & 33.00 & 7.24 & -- \\
& DP-Large
& 16.70 & 15.46 & \textbf{28.62} & 8.29 & 14.00 \\
& \textsc{RoboCode}
& 20.55 & \textbf{14.09} & 38.50
& \textbf{6.08} & \textbf{12.25} \\

\bottomrule
\end{tabular}
}
\end{table*}

\textsc{RoboCode} achieves the highest or tied success rate on all five tasks.
Its average success rate is \(75.6\%\), compared with \(68.8\%\) for ACT,
\(41.6\%\) for DP-Large, and \(18.8\%\) for the compact Diffusion Policy.
The clearest advantage over ACT occurs on LiftPegUpright, where
\textsc{RoboCode} improves success from \(56\%\) to \(80\%\). It also improves
PickCube from \(86\%\) to \(92\%\), while achieving results comparable to ACT
on PokeCube and StackCube and tying ACT at \(98\%\) on PushCube. These results
show that \textsc{RoboCode} remains competitive with a substantially stronger
ACT baseline while providing the additional structure of a shared primitive
action representation.

Increasing the Diffusion Policy capacity produces a consistent improvement in
success rate across all five tasks. DP-Large raises the average success rate
from \(18.8\%\) to \(41.6\%\), with particularly large gains on PickCube,
PushCube, and StackCube. This result indicates that the compact diffusion
model is capacity limited in this setting. Nevertheless, DP-Large remains
below ACT and \textsc{RoboCode} on every task, suggesting that model capacity
alone does not close the performance gap.

The average-return results provide a complementary view of task progress.
\textsc{RoboCode} achieves the highest return on PickCube, PokeCube, and
StackCube, while ACT obtains the highest return on LiftPegUpright and the
compact Diffusion Policy obtains the highest return on PushCube. In particular,
\textsc{RoboCode} and ACT both achieve \(98\%\) success on PushCube despite
having lower average returns than the diffusion baselines. This discrepancy
illustrates that accumulated return is affected by task-specific reward
shaping, episode duration, and termination behavior, and should therefore be
treated as a supplementary metric rather than a monotonic measure of policy
quality.

Steps to success further separate reliability from execution speed.
\textsc{RoboCode} requires the fewest steps on PickCube, PushCube, and
StackCube, while DP-Large is fastest on PokeCube and the compact Diffusion
Policy has the lowest conditional average on LiftPegUpright. These averages
are computed only over successful episodes and should be interpreted together
with success rate; for example, the compact Diffusion Policy succeeds on only
\(14\%\) of LiftPegUpright episodes. Overall, \textsc{RoboCode} provides the
strongest average completion reliability while maintaining competitive
execution efficiency.





\section{Conclusion}
We presented \textsc{RoboCode}, a two-stage framework for learning reusable
primitive action structure from multi-task robot demonstrations. Stage 1
jointly trains an action encoder and a shared primitive action interface that
represents action chunks through soft compositions of learned primitive codes.
Stage 2 freezes this interface and trains task-specific state encoders to map
current observations into the shared latent space. Experiments on five
ManiSkill manipulation tasks show that the learned interface reconstructs
action chunks accurately and exhibits structured code reuse across tasks.
\textsc{RoboCode} achieves the highest or tied success rate across all five
tasks, with an average success rate of \(75.6\%\), compared with \(68.8\%\)
for ACT and \(41.6\%\) for DP-Large. These results support soft-gated primitive
codebooks as an effective and interpretable intermediate action representation
for robot manipulation.

The current evaluation is limited to simulation, leaving open the question of
how the learned primitive interface transfers under real-world perception,
dynamics, contact, and control uncertainty. Moreover, because the task-specific
state encoders are trained by offline imitation learning, closed-loop execution
may encounter state distributions that differ from those in the expert
demonstrations, leading to compounding prediction errors. Future work may
address these sim-to-real and distribution-shift challenges through domain
randomization, interactive data collection, and reinforcement-learning-based
policy refinement.

\section*{Author Contributions}

Xianyue Peng contributed to research conceptualization,
methodology development and implementation,
simulation and real-robot experiments, and manuscript writing.
Iman Soltani contributed to research conceptualization,
methodological guidance, research supervision, and manuscript review.
Devi Amarsaikhan contributed to real-robot data collection
and experimental execution. The corresponding results
will be included in a future revision of the manuscript.

\bibliographystyle{plain}
\bibliography{ref}
\end{document}